\documentclass[conference, 10pt]{IEEEtran}

\usepackage{cite}
\usepackage{url}
\usepackage{graphicx}
\usepackage{color}
\usepackage{placeins}
\usepackage{float}
\usepackage{tabularx,colortbl}
\usepackage{ifthen}

\makeatletter

\newcounter{author}
\renewcommand{\author}[2][]{
   \stepcounter{author}
   \@namedef{author@\theauthor}{#2}
   \@namedef{authorlabel@\theauthor}{#1}
}

\newcounter{address}
\newcommand{\address}[2][]{
   \stepcounter{address}
   \@namedef{address@\theaddress}{#2}
   \@namedef{addresslabel@\theaddress}{#1}
}

\newcommand{\alsep}{and}

\def\newmaketitle{\par%
  \begingroup%
  \normalfont%
  \def\thefootnote{}
  \def\footnotemark{}
  \let\@makefnmark\relax
  \footnotesize
  \footnotesep 0.7\baselineskip
  \normalsize%
  \twocolumn[\thenewmaketitle\@IEEEaftertitletext]%
  \if@IEEEusingpubid
     \enlargethispage{-\@IEEEpubidpullup}%
  \fi
  \endgroup
  \setcounter{footnote}{0}\let\maketitle\relax\let\@maketitle\relax
  \gdef\@thanks{}%
  \let\thanks\relax}

\def\thenewmaketitle{
  \newpage
  \begin{center}%
    \vskip0.2em{\Huge\@IEEEcompsoconly{\sffamily}\@IEEEcompsocconfonly{\normalfont\normalsize\vskip 2\@IEEEnormalsizeunitybaselineskip
   \bfseries\large}\@title\par}\vskip1.0em\par%
    \vspace{1ex}
    \newcounter{c@author}
    \newcounter{c@tmp}
    \ifthenelse{\value{author}=2}{%
      \newcommand{\liand}{ and }}{%
      \newcommand{\liand}{, and }}
    \ifthenelse{\value{address}<2}{%
      \@nameuse{author@1}%
      \stepcounter{c@author}%
      \whiledo{\value{c@author}<\value{author}}{%
        \setcounter{c@tmp}{\value{author}}%
        \addtocounter{c@tmp}{-\value{c@author}}%
        \ifthenelse{\value{c@tmp}=1}{%
          \renewcommand{\alsep}{\liand}}{\renewcommand{\alsep}{, }}%
        \stepcounter{c@author}\alsep \@nameuse{author@\thec@author}}\\%
    }
    {
      \@nameuse{author@1}${}^{(\ref{\@nameuse{authorlabel@1}})}$%
      \stepcounter{c@author}%
      \whiledo{\value{c@author}<\value{author}}{%
      \setcounter{c@tmp}{\value{author}}%
      \addtocounter{c@tmp}{-\value{c@author}}%
      \ifthenelse{\value{c@tmp}=1}{%
        \renewcommand{\alsep}{\liand}}{\renewcommand{\alsep}{, }}%
      \stepcounter{c@author}\alsep \@nameuse{author@\thec@author}%
        ${}^{(\ref{\@nameuse{authorlabel@\thec@author}})}$%
      }
    }
    \vspace{0.2ex}

    \ifthenelse{\value{address}>0}{%
      \ifthenelse{\value{address}=1}{
        {\@nameuse{address@1}}
      }
      {
        \newcounter{c@address}

        \begin{center}
        \whiledo{\value{c@address}<\value{address}}
        {
          \refstepcounter{c@address}
            ${}^{(\thec@address)}$\,%
              \label{\@nameuse{addresslabel@\thec@address}}%
              \@nameuse{address@\thec@address}\\ %
        }
        \end{center}
      } 
    }
    {
      \relax
    }
  \end{center}
}

\makeatother

\title{Real-Time 4D Radar Perception for Robust Human Detection in Harsh Enclosed Environments}

\author[org1]{Zhenan Liu}
\author[org1]{Yaodong Cui}
\author[org1]{Amir Khajepour}
\author[org1]{George Shaker}

\address[org1]{University of Waterloo, 200 University Ave W, Waterloo, ON N2L3G1, Canada}

\begin{document}

\newmaketitle

\begin{abstract}
This paper introduces a novel methodology for generating controlled, multi-level dust concentrations in a highly cluttered environment representative of harsh, enclosed environments, such as underground mines, road tunnels, or collapsed buildings, enabling repeatable mm-wave propagation studies under severe electromagnetic constraints. We also present a new 4D mmWave radar dataset, augmented by camera and LiDAR, illustrating how dust particles and reflective surfaces jointly impact the sensing functionality. To address these challenges, we develop a threshold-based noise filtering framework leveraging key radar parameters (RCS, velocity, azimuth, elevation) to suppress ghost targets and mitigate strong multipath reflections at the raw data level. Building on the filtered point clouds, a cluster-level, rule-based classification pipeline exploits radar semantics—velocity, RCS, and volumetric spread—to achieve reliable, real-time pedestrian detection without extensive domain-specific training. Experimental results confirm that this integrated approach significantly enhances clutter mitigation, detection robustness, and overall system resilience in dust-laden mining environments.
\end{abstract}

\section{Introduction}

Ensuring robust perception for vehicles and mobile machinery in environments with dust, smoke, or other extreme conditions remains a major challenge, since these factors can drastically degrade the performance of optical sensors such as cameras and LiDAR. Although next-generation 4D radar—featuring larger antenna arrays and cascaded chipsets—produces high-resolution range-Doppler-azimuth-elevation data \cite{9760734}, existing datasets predominantly focus on open spaces in adverse weather \cite{apalffy2022} and smoke \cite{paek2023kradar4dradarobject}, overlooking the complexities of dust-laden, high-clutter, and close-proximity settings typical of underground mining. In such confined scenarios, multi-path reflections and significant interference often generate ghost detections, and high dust concentrations can further obscure sensor outputs.

\begin{figure}
    \centering
    \includegraphics[width=0.5\textwidth]{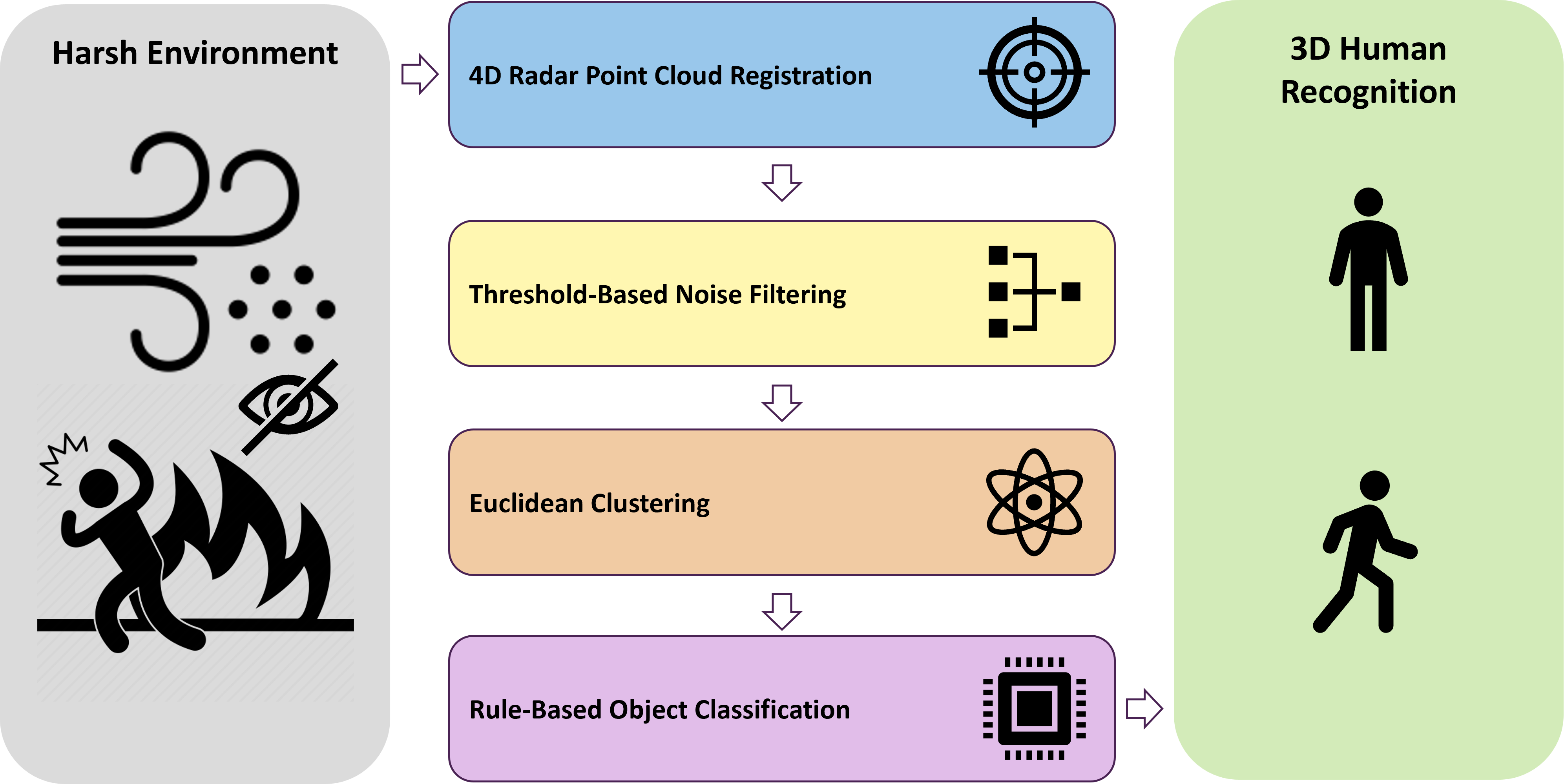}
    \caption{ Framework of a real-time, standalone 4D mmWave
radar perception system for human detection in harsh environment }
    \vspace*{-0.4cm}
    \label{fig:framework}
\end{figure}

This paper introduces a real-time, standalone 4D mmWave radar perception system for human detection in close-spaced, dust-filled indoor settings representative of underground mines and tunnels, fire rescue zones, or collapsed buildings. The proposed system framework is illustrated in \ref{fig:framework}, our contributions include:

\begin{itemize}
    \item A novel methodology for generating controlled multi-level dust concentrations in a closed space enabling systematic and repeatable evaluation of electromagnetic wave propagation at mm-wave frequencies. This methodology is accompanied by a new 4D mmWave radar dataset (together with camera and LiDAR), which reveals how dust particles and reflective surfaces jointly affect the mm-wave channel.

    \item A threshold-based noise filtering framework that exploits key radar parameters (RCS, velocity, azimuth, and elevation) to mitigate strong multipath reflections and suppress ghost targets at the raw data level. By incorporating fundamental mm-wave channel properties, this approach effectively reduces clutter and enhances detection robustness in dust-laden, high-clutter environments.

    \item A cluster-level, rule-based classification pipeline that utilizes radar semantics to reliably detect pedestrians in real time. This design avoids large domain-specific training sets while maintaining high accuracy under challenging electromagnetic propagation conditions caused by both the closed cluttered environment and airborne dust.
\end{itemize}


We validate the proposed system using data from radar, camera, and LiDAR under dusty conditions, demonstrating robust, low-latency human detection and highlighting the system’s potential for safety-critical applications in mining, industrial, and off-road scenarios.

\section{Sensor Setup}

The 4D mmWave radar employed in this work is an Altos imaging radar \cite{altosradar2023}, featuring a four-chip cascaded design with 12 transmit antennas (TX) and 16 receive antennas (RX). This configuration enables the generation of high-density point clouds with an angular resolution of approximately 1.4° in both azimuth and elevation. To evaluate sensor failures and compare registration capabilities under harsh conditions, we also integrate a 40-line LiDAR and an infrared (IR) camera. 

\begin{figure}[!ht]
\vspace*{-0.2cm}
    \centering
    \includegraphics[width=0.45\textwidth]{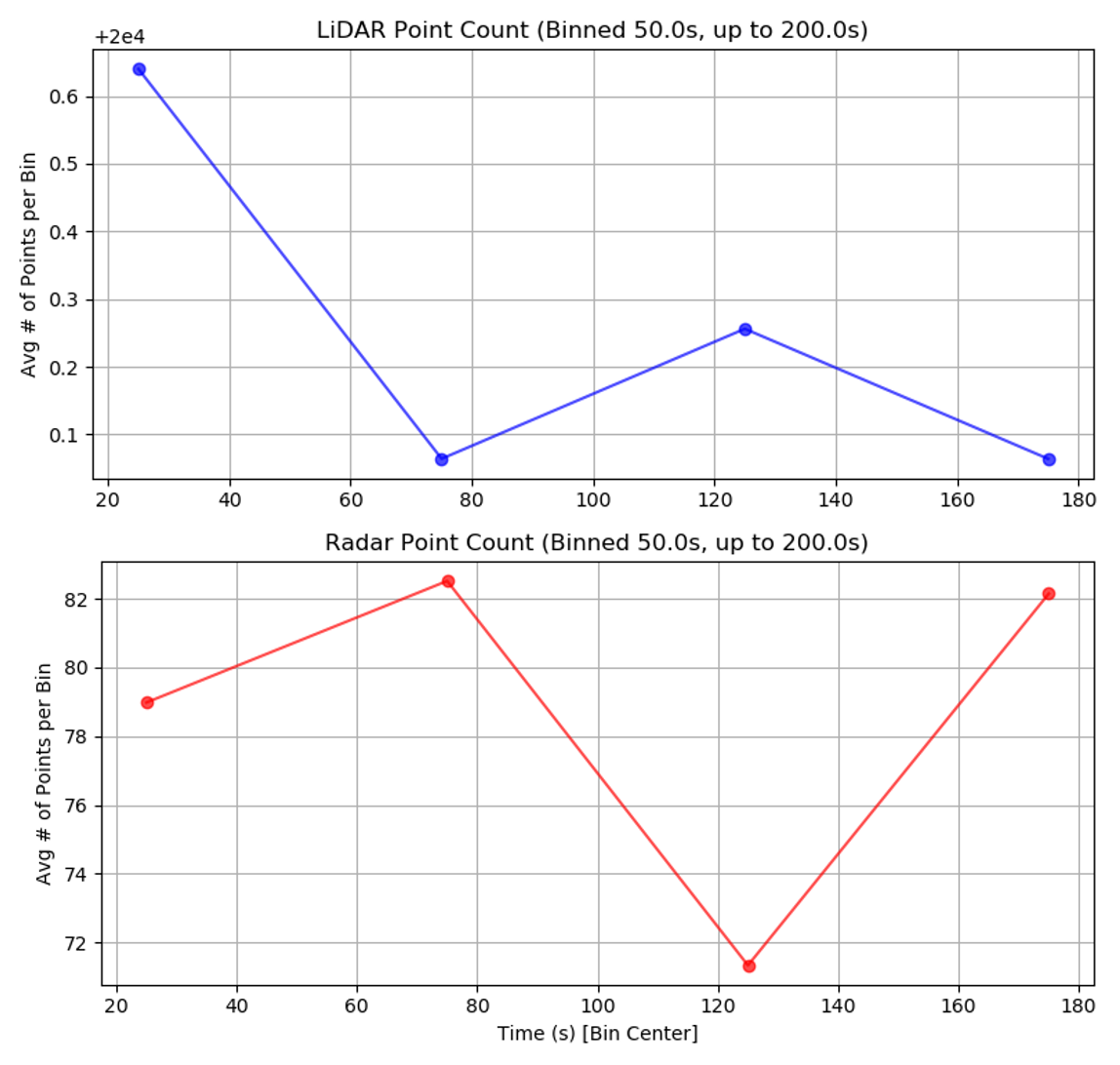}
    \caption{The number of raw point cloud points registered from LiDAR and radar as dust levels rise and the number of pedestrians increases.}
    \label{fig:lidar_radar}
\vspace*{-0.4cm}
\end{figure}

\section{Data Preparation}

To replicate a realistic dusty environment, we positioned a 4D radar, IR camera, and LiDAR inside a $53' \times 10' \times 11'$ truck trailer fitted with metal strips on the ceiling and wooden walls. A sand mixture was dispersed until the dust permeated the entire trailer, while two individuals wearing personal protective equipment walked at various speeds and directions to simulate dynamic scenarios. We collected 9,202 radar point-cloud frames at escalating dust levels. Each detected point is represented as \begin{equation}
    P_i = [x_i, y_i, z_i, RCS_i, v_i, \theta_i, \phi_i]
\end{equation}
where $(x_i, y_i, z_)$ are Cartesian coordinates, $RCS_i$ is the radar cross-section, $v_i$ is point's relative velocity, and $\theta_i$, $\phi_i$ are the azimuth and elevation angles, respectively.
\begin{figure}[!ht]
\vspace*{-0.2cm}
    \centering
    \includegraphics[width=0.45\textwidth]{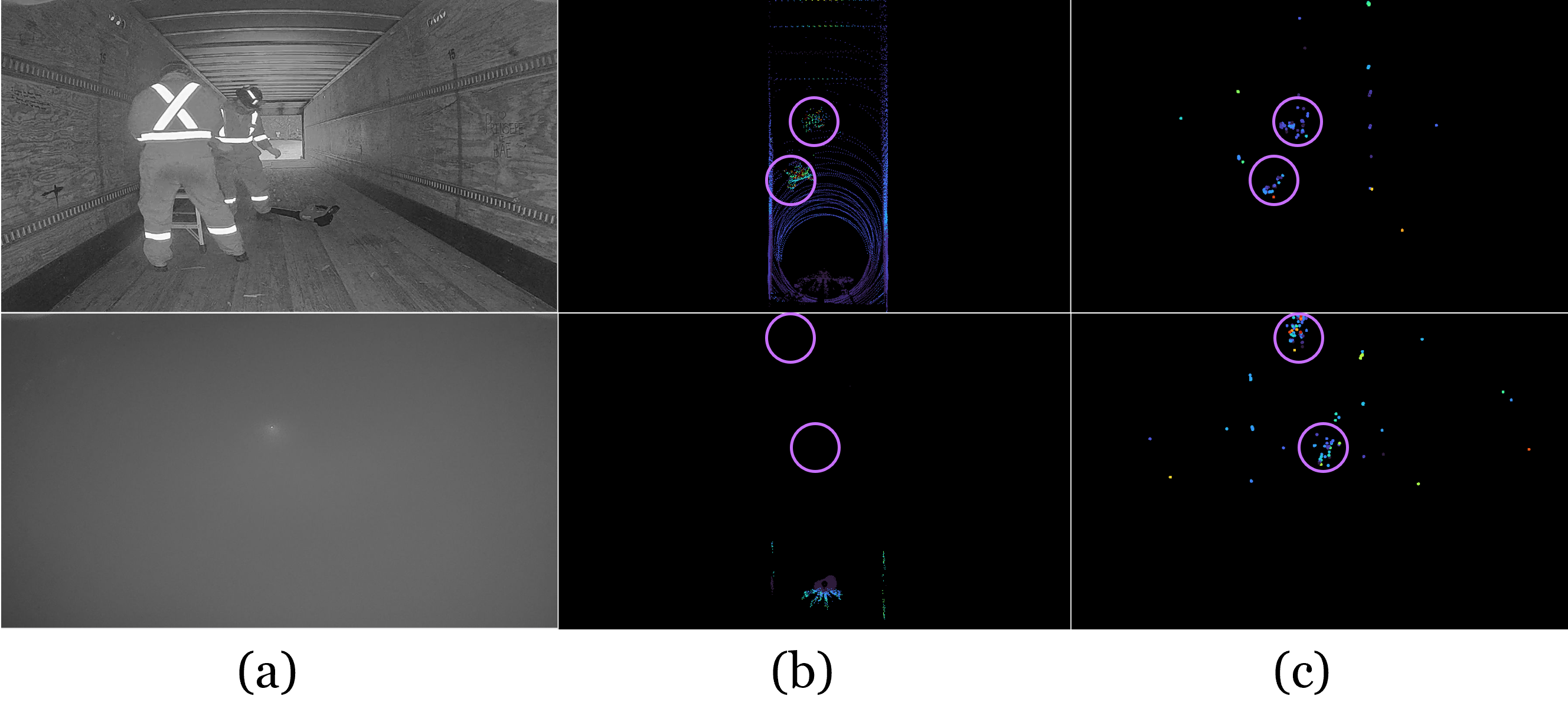}
    \vspace*{-0.4cm}
    \caption{Interference of dust in close-spaced environment to (a) camera, (b) LiDAR and (c) 4D radar. Compared to the circled position of individuals at (b) and (c), while LiDAR lost its detection ability, 4D radar remained consistent.}
    \label{fig:raw_data}
\vspace*{-0.2cm}
\end{figure}

As dust levels increased, the IR camera became blinded, capturing only headlight reflections. Compare to the number of point cloud showed in \ref{fig:lidar_radar},  the LiDAR eventually failed to detect walls or the ceiling, registering merely the reflective labels on PPEs within approximately 1 meter. Despite these severe visibility degradations, the radar consistently generated coherent point clouds at all dust levels, demonstrating its robust performance in cluttered indoor environments.

\section{Radar Noise Filtering}

Close-spaced indoor environments with reflective surfaces pose significant challenges for mmWave radar. Multi-path reflections, generated when signals bounce off walls, ceilings, and other objects, frequently result in ghost targets or distorted echoes. Moreover, highly reflective materials can create unusually large radar cross-section (RCS) values, obscuring real targets. \cite{kato2018autoware} employs threshold-based noise filtering based on range, Azimuth. Benefit from the data enhancement of 4D Radar, we introduce multiple thresholds designed to address various forms of noise:
\begin{itemize}
\item \textbf{RCS Threshold:} Abnormally high or low \(\mathrm{RCS}_i\) values, often caused by strong metal reflectors or negligible signal returns, are excluded. This step significantly reduces the prevalence of ghost targets that originate from reflective interference.

\item \textbf{Azimuth and Elevation Thresholds:} Extreme angles \(\theta_i\) or \(\phi_i\) often represent signal paths that have undergone multiple reflections. By discarding detections whose angles lie outside the physical coverage bounds of the sensor, we limit false positives due to multi-path artifacts.

\item \textbf{Velocity Threshold:} The radar’s ability to measure radial velocity \(v_i\) helps differentiate stationary clutter from moving objects. We distinguish points with velocity magnitudes that are either implausibly high or near zero if they appear at unexpected ranges, thus refining the dataset for subsequent motion-based analyses.
\end{itemize}
The entire thresholding operation runs in $O(n)$ time, as each point is evaluated independently against the specified constraints. This efficiency is crucial for real-time applications, where thousands of points might be processed per frame. By filtering raw data at this early stage, subsequent clustering and classification steps can focus on higher-quality inputs. Consequently, our approach substantially improves accuracy and reduces false detections in dust-laden, cluttered environments, allowing 4D mmWave radar to retain robust performance despite strong interference.

\section{Rule-based Radar Object Classification}\label{IV}

Data-driven 3D object classification remains a significant challenge for mmWave radar due to data sparsity, fluctuating point densities, and radar-specific artifacts. Traditional approaches, such as the model-based 3D classification method in \cite{9319548}, exploit radar cross-section (RCS), velocity, azimuth, and radar heatmaps to categorize both dynamic and static objects. However, these methods typically require domain-specific training datasets to achieve satisfactory accuracy. With the advent of 4D radar, which provides finer angular resolution and more coherent point cloud registration, data-driven classification becomes more feasible—yet the inherent sparsity and measurement noise still pose difficulties.

To circumvent the need for extensive training data and domain adaptation, we employ a \textit{rule-based} 3D object classification pipeline that operates on clusters derived exclusively from 4D radar point clouds. We begin with the filtered set of point cloud $\{P_i\}$, Our clustering procedure is founded on the Euclidean distance $d(P_i, P_j)$ represented as
\begin{equation}
    d(P_i, P_j) = \sqrt{(x_i-x_j)^2+(y_i-y_j)^2+(z_i-z_j)^2} 
\end{equation}
where $P_i,P_j \in P_{filtered}$, and then leverages a KD-tree for efficient nearest-neighbor search,  which allows for logarithmic-time queries. To form clusters, we iterate through each point $P_{i}$ and perform the following:
\begin{enumerate} 
\item \textbf{Neighbor Search:} Use the KD-tree to retrieve all filtered points within a predefined distance $d$ of $P_{i}$
\item \textbf{Connectivity Check:} For each neighbor $P_j$, if it is not already assigned to a cluster, we assign it to $P_{filtered}$'s cluster. 
\item \textbf{Cluster Formation:} Repeat until all points are visited or assigned to a cluster, yielding distinct sets $\{C_1, C_2, ...\}$
\end{enumerate}

\begin{figure}[ht]
\vspace*{-0.2cm}
    \centering
    \includegraphics[width=0.45\textwidth]{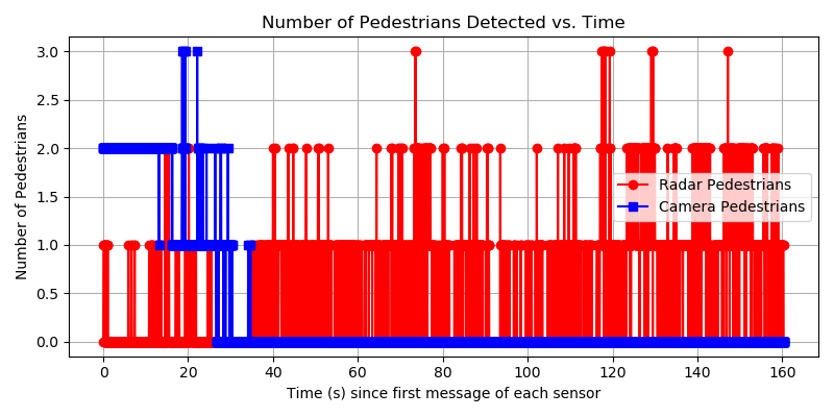}
    \caption{Number of detected pedestrians as dust level increases from (a) YOLOv8 object detection model, (b) standalone radar perception}
    \label{fig:yolo_radar}
\vspace*{-0.4cm}
\end{figure}

Once clusters are formed, we compute \emph{semantic descriptors} per cluster, including \emph{cluster size} ($|C_k|$), \emph{mean velocity} ($\bar{v}_k$), and \emph{mode RCS} ($RCS_{mode,k}$). Using these descriptors, we employ a set of straightforward conditional rules to classify each cluster. For example, a cluster with a relatively small size, moderate mean velocity, and an $RCS_{mode}$ within a known “human” range can be designated as a pedestrian. This label assignment is done without the heavy training requirements or domain-specific tuning that machine learning models typically demand.

We validated this classification strategy by comparing its performance against a YOLOv8 vision-based detection model \cite{Jocher_Ultralytics_YOLO_2023} trained for 100 epochs on the Construction Site Safety dataset \cite{construction-site-safety_dataset}. In dust-free settings, the vision-based model exhibited reliable object detection, but its performance degraded sharply once dust impaired optical visibility as \ref{fig:yolo_radar} illustrated. In contrast, our radar-only approach retained consistent detection and classification across varying dust densities, confirming the robustness of 4D radar perception in real time.


Overall, these results emphasize the potential for 4D mmWave radar in applications where optical sensors fail, while simultaneously acknowledging that deeper data-driven methods may require substantial domain-specific training. By establishing a streamlined rule-based pipeline grounded in KD-tree clustering and semantic thresholds, this work provides a reliable alternative for harsh or visually degraded conditions—without the need for large annotated datasets or computationally intensive learning-based models.

\section{Conclusion}

This paper has presented a real-time, standalone 4D mmWave radar perception system specifically designed to operate in dust-laden, highly cluttered environments, such as those encountered in underground mines, tunnels, and disaster sites. Central to our approach is a novel methodology for generating multiple levels of airborne particulates in a confined setting, accompanied by a new 4D mmWave radar dataset (augmented by camera and LiDAR) that illuminates how dust particles and reflective surfaces jointly impact the radar channel. By integrating a threshold-based noise filtering framework—leveraging critical radar parameters of RCS, velocity, azimuth, and elevation—with a cluster-level, rule-based classification pipeline, we suppress multipath-induced ghost targets at the raw data level and enable reliable real-time pedestrian detection without extensive domain-specific training.

The experimental results demonstrate that high-resolution 4D radar data can significantly bolster detection robustness when conventional optical sensors fail, thus emphasizing the importance of mmWave propagation characteristics and antenna design in harsh industrial applications. Nonetheless, limitations remain in capturing comprehensive environmental context, suggesting the need for temporal data accumulation, object tracking, and sensor fusion strategies to further reduce clutter and refine classification. These ongoing enhancements will deepen our understanding of electromagnetic propagation in dusty, enclosed spaces and expand the applicability of mmWave radar across safety-critical scenarios. Ultimately, by marrying fundamental antenna and propagation insights with advanced signal processing, the proposed system offers a compelling blueprint for next-generation radar-based perception in challenging industrial environments.

\section*{ACKNOWLEDGEMENT}
The authors of this paper would like to acknowledge the support of WSDL, MVSL, Purolator, Cloudhawk, AVRIL, NSERC, MITACS, Rogers, and Google.



%
\bibliographystyle{IEEEtran}
\bibliography{main}

@ARTICLE{9760734,
  author={Bilik, Igal},
  journal={IEEE Intelligent Transportation Systems Magazine}, 
  title={Comparative Analysis of Radar and Lidar Technologies for Automotive Applications}, 
  year={2023},
  volume={15},
  number={1},
  pages={244-269},
  doi={10.1109/MITS.2022.3162886}}

@ARTICLE{apalffy2022,
  author={Palffy, Andras and Pool, Ewoud and Baratam, Srimannarayana and Kooij, Julian F. P. and Gavrila, Dariu M.},
  journal={IEEE Robotics and Automation Letters}, 
  title={Multi-Class Road User Detection With 3+1D Radar in the View-of-Delft Dataset}, 
  year={2022},
  volume={7},
  number={2},
  pages={4961-4968},
  doi={10.1109/LRA.2022.3147324}}

@misc{paek2023kradar4dradarobject,
      title={K-Radar: 4D Radar Object Detection for Autonomous Driving in Various Weather Conditions}, 
      author={Dong-Hee Paek and Seung-Hyun Kong and Kevin Tirta Wijaya},
      year={2023},
      eprint={2206.08171},
      archivePrefix={arXiv},
      primaryClass={cs.CV},
      url={https://arxiv.org/abs/2206.08171}, 
}

@inproceedings{kato2018autoware,
  title={Autoware on board: Enabling autonomous vehicles with embedded systems},
  author={Kato, Shinpei and Tokunaga, Shota and Maruyama, Yuya and Maeda, Seiya and Hirabayashi, Manato and Kitsukawa, Yuki and Monrroy, Abraham and Ando, Tomohito and Fujii, Yusuke and Azumi, Takuya},
  booktitle={Proceedings of the 9th ACM/IEEE International Conference on Cyber-Physical Systems (ICCPS)},
  pages={287--296},
  year={2018},
}

@ARTICLE{9319548,
  author={Cai, Xiuzhang and Giallorenzo, Michael and Sarabandi, Kamal},
  journal={IEEE Transactions on Intelligent Vehicles}, 
  title={Machine Learning-Based Target Classification for MMW Radar in Autonomous Driving}, 
  year={2021},
  volume={6},
  number={4},
  pages={678-689},
  doi={10.1109/TIV.2020.3048944}}

@software{Jocher_Ultralytics_YOLO_2023,
author = {Jocher, Glenn and Qiu, Jing and Chaurasia, Ayush},
license = {AGPL-3.0},
month = jan,
title = {{Ultralytics YOLO}},
url = {https://github.com/ultralytics/ultralytics},
version = {8.0.0},
year = {2023}
}

@misc{
                            construction-site-safety_dataset,
                            title = { Construction Site Safety Dataset },
                            type = { Open Source Dataset },
                            author = { Roboflow Universe Projects },
                            howpublished = { \url{ https://universe.roboflow.com/roboflow-universe-projects/construction-site-safety } },
                            url = { https://universe.roboflow.com/roboflow-universe-projects/construction-site-safety },
                            journal = { Roboflow Universe },
                            publisher = { Roboflow },
                            year = { 2024 },
                            month = { aug },
                            note = { visited on 2025-01-17 },
                            }

@misc{altosradar2023,
  author       = {Altos Radar},
  title        = {Altos Radar Product Page},
  year         = {2023},
  howpublished = {\url{https://www.altosradar.com/product}},
  note         = {Accessed: 2024-09-09}
}

\end{document}